# AN OPTIMIZED RECURSIVE LEARNING ALGORITHM FOR THREE-LAYER FEEDFORWARD NEURAL NETWORKS FOR MIMO NONLINEAR SYSTEM IDENTIFICATIONS

**DAOHANG SHA[1], VLADIMIR B. BAJIC[2]**

*[1]Center for Clinical Epidemiology and Biostatistics*
*University of Pennsylvania School of Medicine*
*525 Blockley Hall*
*423 Guardian Drive*
*Philadelphia, PA 19104, USA*
*E-mail: dsha@mail.med.upenn.edu*

*[2]South African National Bioinformatics Institute (SANBI)*
*University of the Western Cape*
*South Africa*
*E-mail: vlad@sanbi.ac.za*

ABSTRACT—Back-propagation with gradient method is the most popular learning algorithm for feed-forward neural networks. However, it is critical to determine a proper fixed learning rate for the algorithm. In this paper, an optimized recursive algorithm is presented for online learning based on matrix operation and optimization methods analytically, which can avoid the trouble to select a proper learning rate for the gradient method. The proof of weak convergence of the proposed algorithm also is given. Although this approach is proposed for three-layer, feed-forward neural networks, it could be extended to multiple layer feed-forward neural networks. The effectiveness of the proposed algorithms applied to the identification of behavior of a two-input and two-output non-linear dynamic system is demonstrated by simulation experiments.

Key Words: Neural Networks, Learning Algorithms, Back Propagation, Gradient Descent Method

## 1. INTRODUCTION

Due to the approximation to non-linear functions and the capability of self-learning, feed-forward neural networks (FNN) or multilayer perceptions (MLPs) has been widely applied to pattern recognition, function approximation, signal processing, nonlinear system identification and control etc. All these applications are focused on the learning of neural networks. Many approaches have been developed for the learning of neural networks [1]. One of the most popular and widely used learning algorithms for FNN learning is the error back propagation (BP) algorithm [16]. For the applications of the BP algorithm, it is critical to determine a proper fixed learning rate. If the learning rate is large, learning may occur quickly, but it may also become unstable and the FNN system will not learn at all. To ensure stable learning, the learning rate must be sufficiently small. However, with a small learning rate an FNN may be reliably trained, but it may take a long time and thus it can invalidate the purpose of real-time operation. Also, just how small the learning rate should be is unclear. In addition, for different structures of FNN and for different applications, the best fixed learning rates are different. These problems are inherent to the basic learning rule of FNN that relies on the steepest/gradient descent optimization method [24]. The convergence properties of such algorithms are discussed in [11], [7], [19], [5], [21] [25], and [27]. To overcome these problems, many improved BP algorithms are developed such as variable step size, adaptive learning [14] and [20], and others [2], [3], [6], [9], and [23]. There are other ways to accelerate the MLP's learning by the use of second-order gradient based nonlinear optimizing methods, such as the conjugate gradient algorithm (see [3]) or the Levenberg-Marquardt based method (see [7]). The crucial drawbacks of these methods, however, are that in many applications computational demands are so large that their effective use in on-line identification problems is not viable. Also, a common problem with the all above mentioned methods is a non-optimal choice of the learning rate even with the adaptive change of the learning rate. A solution that uses a near-optimal adaptive learning rate is proposed in [10]. Such types of algorithms for identification of single-input-single-output systems were developed in [17] and [18]. In [26], the adaptive learning rates for fuzzy NN based on the Lyapunov function were proposed, which can guarantee the convergence of the tracking error but may not be optimized.

In this paper, we extend the method of [17] and [18] and derive different optimized adaptive learning rate based on the optimization method – extreme value theory from a standard BP algorithm for a three-layer FNN with multiple inputs and multiple outputs used for the on-line identification of multiple input and multiple output (MIMO) non-linear systems. These results allow a near-optimal selection of the learning rate at each iteration step during the learning process. Since FNNs are widely used for the identification of non-linear systems (see [12] and [22]), as well as for prediction of their behavior (see [15]), we have tested the proposed recursive algorithm for the on-line identification of a non-linear system with two inputs and two outputs.

## 2. FEED-FORWARD NEURAL NETWORKS MODEL

Three layer networks with sufficiently many neurons in the hidden layer have been proven to be capable of approximating any Borel measurable function in any accuracy [8]. So a three-layer FNN will be used in this paper, which has $IN$ nodes of the input layer, $HN$ nodes of the hidden layer and $ON$ nodes of the output layer. However, the methodology presented is general and can be applied to FNN consisting of any numbers of hidden layers and neurons. The model of FNNs in matrix form can be written as

$$\hat{y}(t+1)=\begin{bmatrix} f\left[\sum_{i=1}^{HN} s\left(\sum_{j=1}^{IN} w_{ij}(t)x_j(t)+w_{i0}(t)\right)v_{1i}(t)+v_{10}(t)\right] \\ f\left[\sum_{i=1}^{HN} s\left(\sum_{j=1}^{IN} w_{ij}(t)x_j(t)+w_{i0}(t)\right)v_{2i}(t)+v_{20}(t)\right] \\ \cdots \\ f\left[\sum_{i=1}^{HN} s\left(\sum_{j=1}^{IN} w_{ij}(t)x_j(t)+w_{i0}(t)\right)v_{ONi}(t)+v_{ON0}(t)\right] \end{bmatrix}=F\left[V^T(t)S\left(W(t)x(t)\right)\right],$$

where $x=\begin{bmatrix}x_0 & x_1 & \dots & x_{IN}\end{bmatrix}^T \in R^{(IN+1)\times 1}$, $\hat{y}=\begin{bmatrix}\hat{y}_1 & \hat{y}_2 & \dots & \hat{y}_{ON}\end{bmatrix}^T \in R^{ON\times 1}$,

$$V=\begin{bmatrix} v_{10} & v_{20} & \cdots & v_{ON0} \\ v_{11} & v_{21} & \cdots & v_{ON1} \\ \cdots & \cdots & \cdots & \cdots \\ v_{1HN} & v_{2HN} & \cdots & v_{ONHN} \end{bmatrix}=\begin{bmatrix}v_1 & v_2 & \dots & v_{ON}\end{bmatrix}\in R^{(HN+1)\times ON},$$

$v_i=\begin{bmatrix}v_{io} & v_{i1} & \dots & v_{iHN}\end{bmatrix}^T \in R^{(HN+1)\times 1}$, $i=1,2,\dots,ON$,

$$W = \begin{bmatrix} w_{10} & w_{11} & \cdots & w_{1IN} \\ w_{20} & w_{21} & \cdots & w_{2IN} \\ \cdots & \cdots & \cdots & \cdots \\ w_{HN0} & w_{HN1} & \cdots & w_{HNIN} \end{bmatrix} = \begin{bmatrix} w_0 & w_1 & \ldots & w_{IN} \end{bmatrix} \in R^{HN\times(IN+1)},$$

$S(Wx) = \left[s(net_0^H) \quad s(net_1^H) \quad \ldots \quad s(net_{HN}^H)\right]^T \in R^{(HN+1)\times 1}$, $net_i^H = \sum_{j=0}^{IN} w_{ij}(t)x_j(t)$, $i = 0,1,\ldots,HN$, is the output of the $i$ -th hidden node, $x_0(t) \equiv 1$ and $s(net_0^H) \equiv 1$. Here $x_j(t)$, $j = 1,\ldots,IN$, are the inputs of the neural network; $\hat{y}$ is the output of the neural network; $w_{ij}(t)$, $i = 1,\ldots,HN$, $j = 1,\ldots,IN$, are the weights from the input layer to the hidden layer; $w_{i0}(t)$, $i = 1,\ldots,HN$, are the biases of the hidden nodes; $v_{ij}(t)$, $i = 1,\ldots,HN$, $j = 1,\ldots,ON$, are the weights from the hidden layer to the output layer; $v_{i0}(t)$ is the bias of the output node; $t$ is a time index; $s$ is the activation function of the nodes for the hidden layer and $f$ for the output layer. The activation function for non-linear nodes is assumed to be a symmetric hyperbolic tangent function, i.e. $s(x) = \tanh(xu_0^{-1})$, and its derivatives are $s'(x) = u_0^{-1}[1 - s^2(x)]$, $s''(x) = -\frac{2}{u_0^2}s(x)[1 - s^2(x)]$, where $u_0$ is the shape factor of the activation function. In what follows $a^T$ is the transpose of a matrix or a vector $a$ and $a'$ denotes a partial derivative of $a$.

## 3. RECURSIVE LEARNING ALGORITHM

Batch learning of neural networks is highly effective. In batch learning, however, the computation of each learning step is very big, especially when the big sample data sets are presented. Also, the neural networks must be re-learning again as new data become available. And it can not be suitable for the on-line learning as the neural networks are used as the predictor or the filter of nonlinear systems. To overcome these defaults, the learning of neural networks should be iterative, allowing the parameters of neural networks to be updated at each sample interval as new data become available. So in the error squares cost function, a weighting sequence should be chosen to weight higher on more recent data at time $t$. This can be achieved, for example, by using an exponential forgetting mechanism in the error criterion like recursive algorithm with the forgetting factor, i.e.

$$E(t) = \frac{1}{2}\sum_{j=1}^{t}\lambda^{t-j}\sum_{i=1}^{ON}e_i^2(j) = \frac{1}{2}\sum_{j=1}^{t}\lambda^{t-j}e^T(j)e(j) = \frac{1}{2}\sum_{j=1}^{t}\lambda^{t-j}[y(j) - \hat{y}(j)]^T[y(j) - \hat{y}(j)], \quad (1)$$

where $0 < \lambda \le 1$ is the forgetting factor; error vector $e(j) = y(j) - \hat{y}(j)$. Thus, the old measurements in the criterion are exponentially discounted by a factor of $\lambda$ and the newest measurement at time $j = t$ will be fully counted because of $\lambda^{t-j}\big|_{j=t} = 1$. To obtain the standard BP algorithm, let us introduce following two lemmas.

**Lemma 1** $\frac{\partial\left[v^T S(Wx)\right]}{\partial W} = \bar{S}'(Wx)\bar{v}x^T$, *where* $S'(Wx)$ *is the Jacobean matrix of* $S(Wx)$, *i.e.* $\bar{S}'(Wx) = \operatorname{diag}\left[s'(net_1) \quad s'(net_2) \quad \cdots \quad s'(net_{HN})\right]$, $s'(net_i) = \frac{\partial s(net_i)}{\partial net_i}$, $i = 0,\cdots,HN$, $\bar{v} = \left[v_1 \quad \ldots \quad v_{HN}\right]^T \in R^{HN\times 1}$.

**Lemma 2** *The total time derivative of the FNN with single output* $v^T S(Wx)$ *is given by* $\frac{d[v^T S(Wx)]}{dt} = S^T(Wx)\frac{dv}{dt} + \bar{v}^T \bar{S}'(Wx)\frac{dW}{dt}x = \frac{dv^T}{dt}S(Wx) + \bar{v}^T \bar{S}'(Wx)\frac{dW}{dt}x.$

The proofs of these two lemmas see [18]. These two lemmas are in general form for any single output. In the following, we will add the subscript $i$ to represent a specific output properly.

### 3.1 Standard BP with Recursive Learning Formulae

Differentiating equation (1) with respect to $V$ gives

$$\nabla_V E(t) = \frac{\partial E(t)}{\partial V(t)} = \frac{\partial}{\partial V(t)}\left[\frac{1}{2}\sum_{j=1}^{t}\lambda^{t-j}\sum_{i=1}^{ON}e_i^2(j)\right] = \frac{1}{2}\sum_{j=1}^{t}\lambda^{t-j}\sum_{i=1}^{ON}\frac{\partial}{\partial V(t)}e_i^2(j) = \sum_{j=1}^{t}\lambda^{t-j}\sum_{i=1}^{ON}e_i(j)\frac{\partial e_i(j)}{\partial V(j)}$$

$$= \sum_{j=1}^{t}\lambda^{t-j}\sum_{i=1}^{ON}e_i(j)\frac{\partial[y_i(j)-\hat{y}_i(j)]}{\partial V(j)} = -\sum_{j=1}^{t}\lambda^{t-j}\sum_{i=1}^{ON}e_i(j)\frac{\partial \hat{y}_i(j)}{\partial V(j)}$$

$$= -\sum_{j=1}^{t}\lambda^{t-j}\sum_{i=1}^{ON}e_i(j)\begin{bmatrix} \frac{\partial \hat{y}_i(j)}{\partial v_{10}} & \frac{\partial \hat{y}_i(j)}{\partial v_{20}} & \cdots & \frac{\partial \hat{y}_i(j)}{\partial v_{i0}} & \cdots & \frac{\partial \hat{y}_i(j)}{\partial v_{ON0}} \\ \frac{\partial \hat{y}_i(j)}{\partial v_{11}} & \frac{\partial \hat{y}_i(j)}{\partial v_{21}} & \cdots & \frac{\partial \hat{y}_i(j)}{\partial v_{i1}} & \cdots & \frac{\partial \hat{y}_i(j)}{\partial v_{ON1}} \\ \cdots & \cdots & \cdots & \cdots & \cdots & \cdots \\ \frac{\partial \hat{y}_i(j)}{\partial v_{1HN}} & \frac{\partial \hat{y}_i(j)}{\partial v_{2HN}} & \cdots & \frac{\partial \hat{y}_i(j)}{\partial v_{iHN}} & \cdots & \frac{\partial \hat{y}_i(j)}{\partial v_{ONHN}} \end{bmatrix}$$

$$= -\sum_{j=1}^{t}\lambda^{t-j}\sum_{i=1}^{ON}e_i(j)f'[v_i^T S(Wx(j))]\begin{bmatrix} 0 & 0 & \cdots & s_{0,j} & \cdots & 0 \\ 0 & 0 & \cdots & s_{1,j} & \cdots & 0 \\ \cdots & \cdots & \cdots & \cdots & \cdots & \cdots \\ 0 & 0 & \cdots & s_{HN,j} & \cdots & 0 \end{bmatrix}$$

$$= -\sum_{j=1}^{t}\lambda^{t-j}\begin{bmatrix} e_{1,j}f'_{1,j}s_{0,j} & \cdots & e_{i,j}f'_{i,j}s_{0,j} & \cdots & e_{ON,j}f'_{ON,j}s_{0,j} \\ e_{1,j}f'_{1,j}s_{1,j} & \cdots & e_{i,j}f'_{i,j}s_{1,j} & \cdots & e_{ON,j}f'_{ON,j}s_{1,j} \\ \cdots & \cdots & \cdots & \cdots & \cdots \\ e_{1,j}f'_{1,j}s_{HN,j} & \cdots & e_{i,j}f'_{i,j}s_{HN,j} & \cdots & e_{ON,j}f'_{ON,j}s_{HN,j} \end{bmatrix}$$

$$= -\sum_{j=1}^{t}\lambda^{t-j}\begin{bmatrix} s_{0,j} \\ s_{1,j} \\ \cdots \\ s_{HN,j} \end{bmatrix}\begin{bmatrix} e_{1,j}f'_{1,j} & e_{2,j}f'_{2,j} & \cdots & e_{ON,j}f'_{ON,j} \end{bmatrix}$$

$$= -\sum_{j=1}^{t}\lambda^{t-j}S(j)\begin{bmatrix} e_{1,j} & e_{2,j} & \cdots & e_{ON,j} \end{bmatrix}\begin{bmatrix} f'_{1,j} & 0 & \cdots & 0 \\ 0 & f'_{2,j} & \cdots & 0 \\ \cdots & \cdots & \cdots & \cdots \\ 0 & 0 & \cdots & f'_{ON,j} \end{bmatrix}$$

$$= -\sum_{j=1}^{t}\lambda^{t-j}S(j)e^T(j)F'(j) = \lambda\nabla_V E(t-1) - S(t)e^T(t)F'(t).$$

Notice that some short notations are made here for simplicity, e.g. $e_i(j) = e_{i,j}$, $f'[v_i^T S(Wx(j))] = f'_{i,j}$, etc.

Similar to the above procedure, differentiating equation (1) with respect to $W$ gives

$$\nabla_W E(t) = \frac{\partial E(t)}{\partial W(t)} = \sum_{j=1}^{t} \lambda^{t-j} \sum_{i=1}^{ON} e_i(j) \frac{\partial e_i(j)}{\partial W(j)} = -\sum_{j=1}^{t} \lambda^{t-j} \sum_{i=1}^{ON} e_i(j) \frac{\partial \hat{y}_i(j)}{\partial W(j)}$$

$$= -\sum_{j=1}^{t} \lambda^{t-j} \sum_{i=1}^{ON} e_i(j) f'[v_i^T S(Wx(j))] \frac{\partial [v_i^T S(Wx(j))]}{\partial W}, \text{ (apply Lemma 1)}$$

$$= -\sum_{j=1}^{t} \lambda^{t-j} \sum_{i=1}^{ON} e_{i,j} f'_{i,j} \bar{S}'_j \bar{v}_i x_j^T = -\sum_{j=1}^{t} \lambda^{t-j} \bar{S}'_j \left\{ \sum_{i=1}^{ON} e_{i,j} f'_{i,j} \bar{v}_i \right\} x_j^T$$

$$= -\sum_{j=1}^{t} \lambda^{t-j} \bar{S}'_j \begin{bmatrix} e_{1,j} f'_{1,j} v_{11} + e_{2,j} f'_{2,j} v_{21} + \ldots + e_{ON,j} f'_{ON,j} v_{ON1} \\ e_{1,j} f'_{1,j} v_{12} + e_{2,j} f'_{2,j} v_{22} + \ldots + e_{ON,j} f'_{ON,j} v_{ON2} \\ \ldots \\ e_{1,j} f'_{1,j} v_{1HN} + e_{2,j} f'_{2,j} v_{2HN} + \ldots + e_{ON,j} f'_{ON,j} v_{ONHN} \end{bmatrix} x_j^T$$

$$= -\sum_{j=1}^{t} \lambda^{t-j} \bar{S}'_j \begin{bmatrix} v_{11} & v_{21} & \ldots & v_{ON1} \\ v_{12} & v_{22} & \ldots & v_{ON2} \\ \ldots & \ldots & \ldots & \ldots \\ v_{1HN} & v_{2HN} & \ldots & v_{ONHN} \end{bmatrix} \begin{bmatrix} e_{1,j} f'_{1,j} \\ e_{2,j} f'_{2,j} \\ \ldots \\ e_{ON,j} f'_{ON,j} \end{bmatrix} x_j^T$$

$$= -\sum_{j=1}^{t} \lambda^{t-j} \bar{S}'_j \begin{bmatrix} v_{11} & v_{21} & \ldots & v_{ON1} \\ v_{12} & v_{22} & \ldots & v_{ON2} \\ \ldots & \ldots & \ldots & \ldots \\ v_{1HN} & v_{2HN} & \ldots & v_{ONHN} \end{bmatrix} \begin{bmatrix} f'_{1,j} & 0 & \ldots & 0 \\ 0 & f'_{2,j} & \ldots & 0 \\ \ldots & \ldots & \ldots & \ldots \\ 0 & 0 & \ldots & f'_{ON,j} \end{bmatrix} \begin{bmatrix} e_{1,j} \\ e_{2,j} \\ \ldots \\ e_{ON,j} \end{bmatrix} x_j^T$$

$$= -\sum_{j=1}^{t} \lambda^{t-j} \bar{S}'(j) \bar{V}(j) F'(j) e(j) x^T(j) = \lambda \nabla_W E(t-1) - \bar{S}'(t) \bar{V}(t) F'(t) e(t) x^T(t),$$

where $\bar{S}'_j = \text{diag}\left[ s'_{1j} \;\; \ldots \;\; s'_{HNj} \right] \in R^{HN \times HN}, \;\; j = 1,\ldots,t,$

$$s'_{i,j} = s'(net^H_{i,j}) = \frac{\partial s(net^H_{i,j})}{\partial net^H_{i,j}}, \; i = 1,\ldots,HN, \;\; j = 1,\ldots,t,$$

$$\bar{V} = \begin{bmatrix} v_{11} & v_{21} & \ldots & v_{ON1} \\ v_{12} & v_{22} & \ldots & v_{ON2} \\ \ldots & \ldots & \ldots & \ldots \\ v_{1HN} & v_{2HN} & \ldots & v_{ONHN} \end{bmatrix} = \left[ \bar{v}_1 \;\; \bar{v}_2 \;\; \ldots \;\; \bar{v}_{ON} \right] \in R^{HN \times ON},$$

$\bar{v}_i = \left[ v_{i1} \;\; \ldots \;\; v_{iHN} \right]^T \in R^{HN \times 1}, \; i = 1,\ldots,HN, \; F'_j = \text{diag}\left[ f'_{1,j} \;\; f'_{2,j} \;\; \ldots \;\; f'_{ON,j} \right] \in R^{ON \times ON}, \;\; j = 1,\ldots,t,$

$$f'_{i,j} = f'[v_i^T S(Wx_j)] = \frac{\partial f[v_i^T S(Wx_j)]}{\partial [v_i^T S(Wx_j)]}, \; i = 1,\ldots,ON, \;\; j = 1,\ldots,t.$$

Therefore, by applying the steepest descent method, the recursive updating of parameter $V$ and $W$ are given by, respectively

$$\Delta V(t) = -\eta \nabla_V E(t), \;\; \Delta W(t) = -\eta \nabla_W E(t), \tag{2, 3}$$

where $\nabla_V E(t) = \lambda \nabla_V E(t-1) - S(t) e^T(t) F'(t), \;\; \nabla_W E(t) = \lambda \nabla_W E(t-1) - \bar{S}'(t) \bar{V}(t) F'(t) e(t) x^T(t),$ (4, 5)

and $\eta$ is a constant learning rate for standard BP algorithm. In the following section, we will derive the optimized learning rate based on the analysis of training error.

### *3.2 Optimized Learning Rate for Recursive Learning*

By applying Lemma 2, together with equations (2) and (3) gives the change in the output of neural networks

$$\Delta\hat{y}(t+1) = \Delta F[V^T S(Wx)] = \begin{bmatrix} \Delta f_1 \\ \Delta f_2 \\ \cdots \\ \Delta f_{ON} \end{bmatrix} \approx \begin{bmatrix} f_1' \cdot \left(S^T \Delta v_1 + \bar{v}_1^T \bar{S}' \Delta Wx\right) \\ f_2' \cdot \left(S^T \Delta v_2 + \bar{v}_2^T \bar{S}' \Delta Wx\right) \\ \cdots \\ f_{ON}' \cdot \left(S^T \Delta v_{ON} + \bar{v}_{ON}^T \bar{S}' \Delta Wx\right) \end{bmatrix}$$

$$= \begin{bmatrix} f_1' \cdot \left(S^T\left[-\eta \nabla_{v_1} E(t)\right] + \bar{v}_1^T \bar{S}' \left[-\eta \nabla_W E(t)\right] x\right) \\ f_2' \cdot \left(S^T\left[-\eta \nabla_{v_2} E(t)\right] + \bar{v}_2^T \bar{S}' \left[-\eta \nabla_W E(t)\right] x\right) \\ \cdots \\ f_{ON}' \cdot \left(S^T\left[-\eta \nabla_{v_{ON}} E(t)\right] + \bar{v}_{ON}^T \bar{S}' \left[-\eta \nabla_W E(t)\right] x\right) \end{bmatrix} = -\eta \begin{bmatrix} f_1' \cdot \left(S^T \nabla_{v_1} E(t) + \bar{v}_1^T \bar{S}' \nabla_W E(t) x\right) \\ f_2' \cdot \left(S^T \nabla_{v_2} E(t) + \bar{v}_2^T \bar{S}' \nabla_W E(t) x\right) \\ \cdots \\ f_{ON}' \cdot \left(S^T \nabla_{v_{ON}} E(t) + \bar{v}_{ON}^T \bar{S}' \nabla_W E(t) x\right) \end{bmatrix}$$

$$= -\eta \begin{bmatrix} f_1' & 0 & \cdots & 0 \\ 0 & f_2' & \cdots & 0 \\ \cdots & \cdots & \cdots & \cdots \\ 0 & 0 & \cdots & f_{ON}' \end{bmatrix} \begin{bmatrix} S^T\left[\lambda \nabla_{v_1} E(t-1) - e_1 f_1' S\right] + \bar{v}_1^T \bar{S}' \nabla_W E(t) x \\ S^T\left[\lambda \nabla_{v_2} E(t-1) - e_2 f_2' S\right] + \bar{v}_2^T \bar{S}' \nabla_W E(t) x \\ \cdots \\ S^T\left[\lambda \nabla_{v_{ON}} E(t-1) - e_{ON} f_{ON}' S\right] + \bar{v}_{ON}^T \bar{S}' \nabla_W E(t) x \end{bmatrix}$$

$$= -\eta F'\left[\lambda \nabla_V^T E(t-1) S - F' e S^T S + \bar{V}^T \bar{S}' \left(\lambda \nabla_W E(t-1) - \bar{S}' \bar{V} F' e x^T\right) x\right]$$

$$= -\eta\lambda F'\left[\nabla_V^T E(t-1) S + \bar{V}^T \bar{S}' \nabla_W E(t-1) x\right] + \eta F'\left(F' e S^T S + \bar{V}^T \bar{S}' \bar{S}' \bar{V} F' e x^T x\right)$$

$$= -\eta\lambda F'\left[\nabla_V^T E(t-1) S + \bar{V}^T \bar{S}' \nabla_W E(t-1) x\right] + \eta F'\left(S^T S I_{ON^2} + \bar{V}^T \bar{S}' \bar{S}' \bar{V} x^T x\right) F' e$$

$$= -\eta\lambda\varepsilon(t-1) + \eta\zeta(t) e(t),$$

where $\varepsilon(t-1) = F'\left[\nabla_V^T E(t-1) S + \bar{V}^T \bar{S}' \nabla_W E(t-1) x\right]$, $\zeta(t) = F'\left(S^T S I_{ON^2} + \bar{V}^T \bar{S}' \bar{S}' \bar{V} x^T x\right) F'$. Finally, the change of the output of neural networks is $\Delta\hat{y}(t+1) \approx -\eta\lambda\varepsilon(t-1) + \eta\zeta(t) e(t)$.

Assuming that $|\Delta y(t+1)| \ll |\Delta\hat{y}(t+1)|$ i.e. that the absolute value of the change of the process output is much smaller than the absolute value of the change of the NN output. This implies that the change of the process output can be ignored comparing with the change of neural network output during the learning process [18]. Usually, we can increase the number of nodes of hidden layer of neural networks to make this assumption be true. Now, let us consider the tracking error between the output of the plant and the output of neural networks

$$e(t+1) = e(t) + \Delta y(t+1) - \Delta\hat{y}(t+1) \approx e(t) + \eta\lambda\varepsilon(t-1) - \eta\zeta(t) e(t). \qquad (6)$$

The objective is to derive an optimal learning rate $\eta$ at iteration $t$ at which some cost function will be minimized. To do that, let us define a new cost function as $E_2(t+1) = 0.5 e^T(t+1) e(t+1)$. Insert the tracking error equation (6) into the new cost function, we get $E_2(t+1) = 0.5\left[e(t) - \eta\zeta(t) e(t) + \eta\lambda\varepsilon(t-1)\right]^T \left[e(t) - \eta\zeta(t) e(t) + \eta\lambda\varepsilon(t-1)\right]$, which gives $E_2(t+1)$ as a function of the learning rate $\eta$. To derive the optimized learning rate that minimize $E_2(t+1)$, we use the following conditions of extreme values

$$\left.\frac{dE_2(t+1)}{d\eta}\right|_{\eta=\eta^*(t)} = -0.5[\zeta(t)e(t)-\lambda\varepsilon(t-1)]^T\left[e(t)-\eta^*(t)\zeta(t)e(t)+\eta^*(t)\lambda\varepsilon(t-1)\right]$$

$$-0.5\left[e(t)-\eta^*(t)\zeta(t)e(t)+\eta^*(t)\lambda\varepsilon(t-1)\right]^T[\zeta(t)e(t)-\lambda\varepsilon(t-1)]=0,$$

$$\left.\frac{d^2E_2(t+1)}{d\eta^2}\right|_{\eta=\eta^*(t)} = [\zeta(t)e(t)-\lambda\varepsilon(t-1)]^T[\zeta(t)e(t)-\lambda\varepsilon(t-1)]>0.$$

Since the second order derivative is positive, the optimum value of the learning rate which will minimize $E_2(t+1)$ is found from the first condition to be

$$\eta^*(t)=\frac{[\zeta(t)e(t)-\lambda\varepsilon(t-1)]^T e(t)}{[\zeta(t)e(t)-\lambda\varepsilon(t-1)]^T[\zeta(t)e(t)-\lambda\varepsilon(t-1)]}. \quad (7)$$

The above formula shows that the optimal learning rate is various and only dependent on the current and previous states and current output errors of neural networks, but not on the initial weights. Thus ones don't need to worry about selecting a proper fixed learning rate for the standard BP algorithm with different structure of neural networks such as with different initial weights and different numbers of neurons in neural networks.

### *3.3 Recursive Algorithm with Forgetting Factor is Equivalent to On-line Learning Algorithm with Various Momentum*

Substitute optimal learning rate (7) to (2), and (3), we have $\Delta V(t)=-\eta^*(t)\nabla_V E(t)$ and $\Delta W(t)=-\eta^*(t)\nabla_W E(t)$. Thus, from above equations, ones have the following results immediately, $\nabla_V E(t-1)=-\frac{1}{\eta^*(t-1)}\Delta V(t-1)$ and $\nabla_W E(t-1)=-\frac{1}{\eta^*(t-1)}\Delta W(t-1)$. From (4) and (5), we have $\nabla_V E(t)=-\frac{1}{\eta^*(t-1)}\lambda\Delta V(t-1)-S(t)e^T(t)F'(t)$, $\nabla_W E(t) =-\frac{1}{\eta^*(t-1)}\lambda\Delta W(t-1)-\bar{S}'(t)\bar{V}(t)F'(t)e(t)x^T(t)$. Finally,

$$\Delta V(t) =\eta^*(t)S(t)e^T(t)F'(t)+\frac{\eta^*(t)}{\eta^*(t-1)}\lambda\Delta V(t-1),$$

$$\Delta W(t)=\eta^*(t)\bar{S}'(t)\bar{V}(t)F'(t)e(t)x^T(t)+\frac{\eta^*(t)}{\eta^*(t-1)}\lambda\Delta W(t-1).$$

Here the form of the above equations is exactly same as the standard on-line learning algorithm with various momentums. The first terms on the right hand side of the equations are pure gradient descent methods while the second terms on the right hand side of the equations are the momentum terms. The coefficients for momentum terms are the same both for $\Delta V(t-1)$ and $\Delta W(t-1)$, i.e. $\frac{\eta^*(t)}{\eta^*(t-1)}\lambda$. Like the momentums, the proposed algorithm with the forgetting factor will smooth the updates of weights in neural networks compared to the pure gradient descent methods.

### 3.4 Convergence of New Recursive Learning Algorithm with Optimized Learning Rate

Before discussing the convergence of the new recursive learning algorithm with optimized learning rate, let us first introduce some propositions and corollaries.

**Proposition 1** If $A \in R^{m\times n}$, then $\|A\|_2 = \sqrt{\max_{\mu\in\lambda(A^TA)}\mu}$, i.e. $\|A\|_2$ is the square root of the largest eigenvalue of $A^TA$.

**Corollary 1** If matrix $A \in R^{n\times n}$ is symmetric, then $\|A\|_2 = \max_{\mu\in\lambda(A)}\mu$.

**Proposition 2** If $A \in R^{n\times n}$ is idempotent, i.e. $A^2 = A$, then $\lambda \in \lambda(A) = \{1, 0\}$ , i.e. eigenvalues of idempotent matrices is 1 or 0.

**Corollary 2** If $A \in R^{n\times n}$ is symmetric and idempotent, then $\|A\|_2 = 1$.

Now we will show the convergence of the new recursive learning algorithm, Substitute the optimal learning rate (7) into Equation (6), we have

$$e(t+1) \approx e(t) - \eta^*(t)[\zeta(t)e(t) - \lambda\varepsilon(t-1)] = e(t) - \frac{[\zeta(t)e(t)-\lambda\varepsilon(t-1)][\zeta(t)e(t)-\lambda\varepsilon(t-1)]^T e(t)}{[\zeta(t)e(t)-\lambda\varepsilon(t-1)]^T[\zeta(t)e(t)-\lambda\varepsilon(t-1)]}.$$

Let $x = \zeta(t)e(t) - \lambda\varepsilon(t-1)$, and $M = I - \frac{xx^T}{x^Tx}$, then $e(t+1) \approx M(t)e(t)$. Notice that the error cost function

$$E_2(t+1) = 0.5e^T(t+1)e(t+1) = 0.5\|e(t+1)\|_2^2 \approx 0.5\|M(t)e(t)\|_2^2 \le 0.5\|M(t)\|_2^2\|e(t)\|_2^2 = \|M(t)\|_2^2 E_2(t).$$

It is easy to prove that the matrix $M$ is idempotent and symmetric for $\forall x \ne 0$ , where $x$ is a vector, i.e. $M^2 = MM = \left(I - \frac{xx^T}{x^Tx}\right)\left(I - \frac{xx^T}{x^Tx}\right) = I - 2\frac{xx^T}{x^Tx} + \frac{xx^T}{x^Tx}\frac{xx^T}{x^Tx} = I - \frac{xx^T}{x^Tx} = M$; and $M^T = \left(I - \frac{xx^T}{x^Tx}\right)^T = I - \frac{xx^T}{x^Tx} = M$, i.e. $.M^2 = M$ and $M^T = M$. Apply Corollary 2, we have $\|M(t)\|_2^2 = 1$. So $E_2(t+1) \le E_2(t)$, i.e. the sequence $\{E_2(t)\}$ monotonically decreases as the number of iterative increases. Since $E_2(t)$ is nonnegative, it must converge to some value $E^* \ge 0$, i.e. $\lim_{t\to\infty} E_2(t) = E^*$. Up to now, we have proved the following weak convergence results for recursive learning algorithm with optimal learning rate.

**Theorem** For three-layer FNN $\hat{y}(t+1) = F[V^T(t)S(W(t)x(t))]$ with an error cost function as $E(t) = \frac{1}{2}\Sigma_{j=1}^t \lambda^{t-j}e^T(j)e(j)$ , if using iteration process (2) and (3) with optimal learning rate (7), ones have $E_2(t+1) \le E_2(t), t = 0,1,...$ and $\lim_{t\to\infty} E_2(t) = E^* \ge 0$.

Comparing with the proof of the weak convergence for an on-line gradient method with variable step size (the learning rate goes to zero as the learning step goes to infinity in a fashion $O(1/t)$ for three-layer FNNs in [25], our proof procedure presented here is very concise. Our proof method has also eliminated all those requirements in [27] and [21], such as two-layer FNN (no hidden layer) under the requirements of the outputs of FNN and their first and second order derivatives being uniformly bounded [27].

### 3.5 Variable Forgetting Factors

There are two situations in which a smaller forgetting factor should be used.

- **Start-up forgetting factor**

Due to $\varepsilon(t-1)$ might be very big at the beginning of learning, the optimized learning rate may be negative. So the forgetting factor should initially be small. In the other side, to increase the generalization of NN after finishing learning, the forgetting factor should be big towards unity. Therefore, a variable forgetting factor which does this is given by $\lambda_1(t) = \alpha\lambda_1(t-1) + 1 - \alpha$, where $\alpha = \exp(-1/\tau_f)$, $\tau_f$ is the time constant of the forgetting factor determining how fast $\lambda_1(t)$ changes.

- **Adaptive forgetting factor**

During the learning process of NN, if the prediction error $e(t)$ grows, it may mean that the parameters of systems to be identified changed. The NN model is incorrect and needs adjustment. So we should decrease the forgetting factor and allow NN model to adapt. A variable forgetting factor which allows this is $\lambda_2(t) = \dfrac{s_f(t-1)}{s_f(t)}$, where $s_f(t)$ is a weighted average of the past values of $e^T e$, it is calculated by $s_f(t) = \dfrac{\tau_f - 1}{\tau_f} s_f(t-1) + \dfrac{e^T e}{\tau_f}$, $\tau_f$ again determines the rate of adaptation of $\lambda_2(t)$. Finally, it is possible to combine the start-up forgetting factor $\lambda_1(t)$ with the adaptive forgetting factor $\lambda_2(t)$ to give the variable forgetting factor $\lambda(t)$, thus $\lambda(t) = \lambda_1(t)\lambda_2(t)$.

## 4. SIMULATION RESULTS

For comparison of various learning algorithms, the sum of running mean squared error (RMSE) is computed during the learning procedure using the following formula $RMSE = \dfrac{1}{t+1}\sum_{j=0}^{t}\sum_{i=1}^{ON}\left[y_i(j) - y_i^N(j)\right]^2$. Testing of the expressions for the adaptive optimized change of the learning rate in an on-line fashion is carried out by applying these learning rates in an experiment of on-line identification of a MIMO non-linear dynamic system model borrowed from [4], [13] and the one-step-ahead prediction of system outputs. The different is that we have considered the output measurement noise of the plant during the learning process of neural networks. The structure of the identification/prediction system is depicted in Figure 1. The TDL block denotes the tapped delay line block. The two-input-two-output plant is described by

$$\begin{cases} y_1(t+1) = \dfrac{0.8y_1^3(t) + u_1^2(t)u_2(t)}{2 + y_2^2(t)} + \omega_1(t), \\ y_2(t+1) = \dfrac{y_1(t) - y_1(t)y_2(t) + [u_1(t) - 0.5][u_2(t) + 0.8]}{2 + y_2^2(t)} + \omega_2(t). \end{cases}$$

The input for training the ANN model is $x(t) = [u_1(t)\ u_2(t)\ y_1(t)\ y_2(t)]^T$, and the output (target) vector is $y(t) = [y_1(t+1)\ y_2(t+1)]^T$. Both hidden neurons and output neurons are non-linear. Vector $\omega(t) = [\omega_1(t)\ \omega_2(t)]^T$ is measurement noise such as band-limited white noise. The power of noise is 0.001. The inputs applied are $u_1(t) = \sin(2\pi t/250)$ and $u_2(t) = \cos(2\pi t/250)$. The on-line identification of the system model and one-step-ahead prediction of its outputs by such models was carried out using standard BP with different learning rate, as well as with the optimized adaptive learning rates proposed in the paper. To compare with other adaptive learning rate, we also adopted one from [25], i.e. the adaptive learning rate is $\eta(t) = \frac{\eta_0}{1 + t\beta\eta_0}$, where $\eta_0$ is the initial learning rate and $\beta$ is a constant. Both $\eta_0$ and $\beta$ need to be selected and tuned manually for different structure NN before running the simulations. The RMSE during the first 5,000 iteration steps are plotted in Figures 2, 3 and 4. Figure 2 depicts the RMSEs of the standard BP algorithm with the fixed learning rates ($\eta = 0.1,\ 0.4$ and 1.4) as well as with the recursive optimized adaptive learning with 10 hidden neurons. Figure 3 depicts the RMSEs of the standard BP algorithm with the fixed learning rates ($\eta = 0.05,\ 0.35$ and 1.0), as well as with the recursive optimized adaptive learning rate with 20 hidden neurons. Figure 4 depicts the RMSEs of the standard BP algorithm with the fixed learning rates ($\eta = 0.05,\ 0.5$ and 1.0) as well as with the recursive optimized adaptive learning with 20 hidden neurons and with different initial weights. One observes that in this example the different structure and initial weights may need a different optimal learning rate, for example, in Figure 2 (the optimal learning rate $\eta_{opt}$ is around 0.4), in Figure 3 ($\eta_{opt} \sim 0.35$) and in Figure 4 ($\eta_{opt} \sim 0.5$). The optimized learning algorithm, however, always performs the best behavior near the optimal fixed learning rate. Figure 5 shows the input/output signals of the system and neural networks and the changes in learning rate as time changes for the simulation shown in Figure 4. It is clear that the value of learning rate is positive and its range of changes is quite large (from 0.2-4) during entire learning process. The outputs of neural networks follow the outputs of the system very well. The optimal learning rate appears periodically since the outputs of the system are periodical. These results effectively show that the algorithms for the adaptive selection of near-optimal values of $\eta$ are suitable for on-line applications, having the main advantage of automatic selection of the suitable learning rate. There is nothing need to be tuned with the proposed optimized adaptive learning rate for all simulations. However, the computation of the proposed learning rate is much more complex than the adaptive learning rate. Moreover the complexity will increase as the numbers of layers and nodes of neural networks increase.

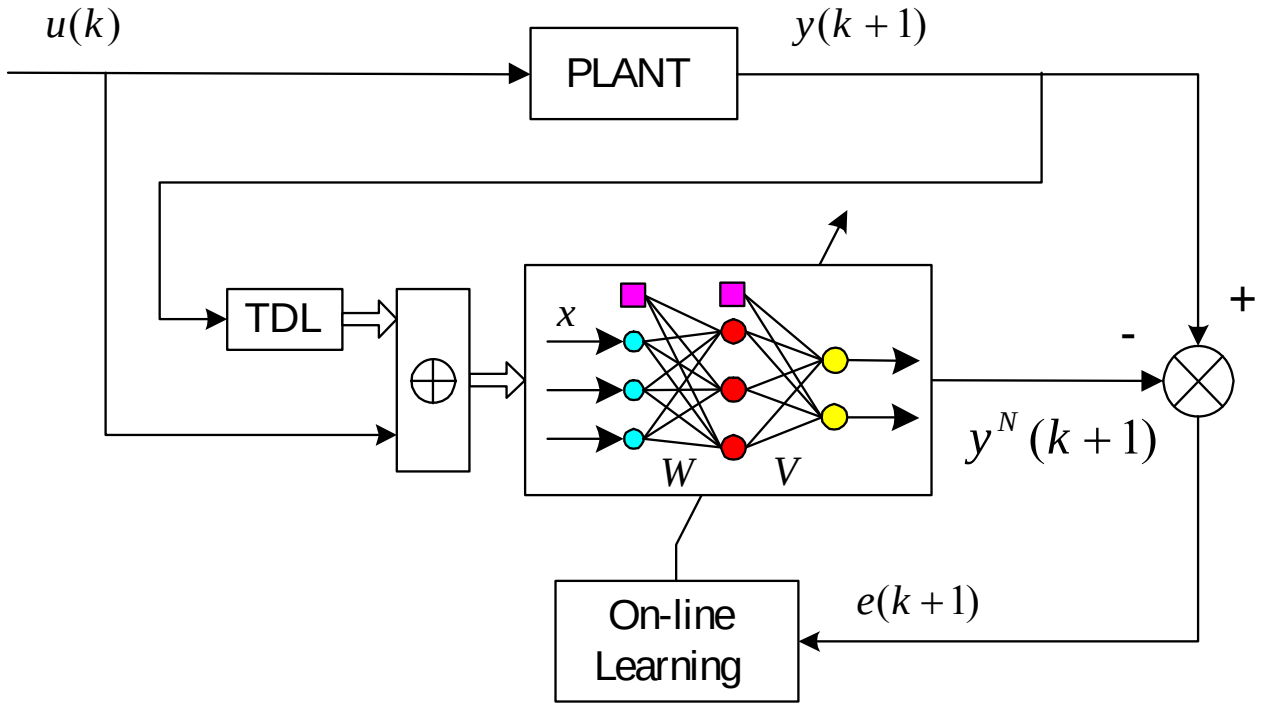

**Figure 1. Diagram of neural networks for on-line identification and prediction of nonlinear system outputs.**

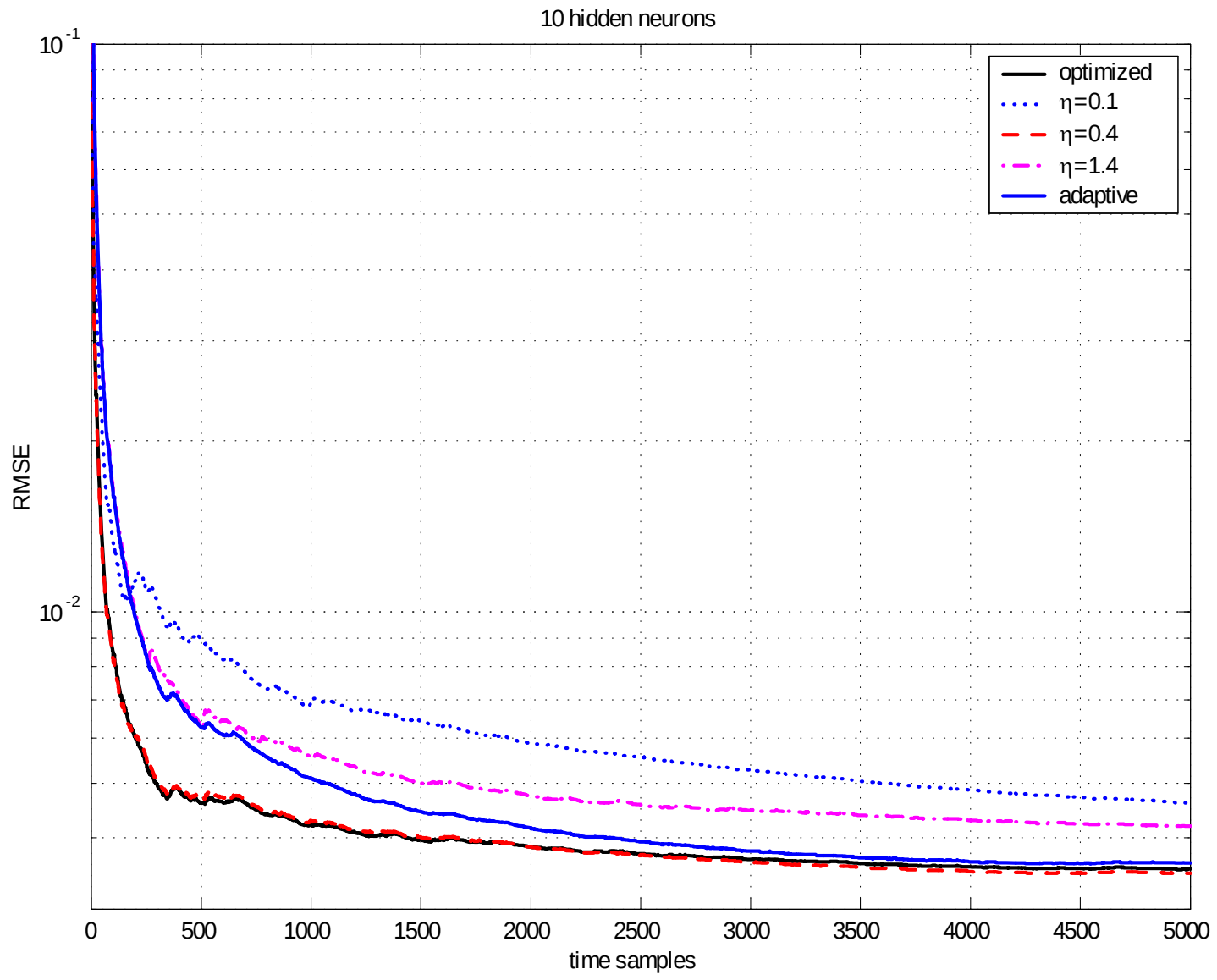

**Figure 2. RMSE in one-step-ahead prediction of system outputs based on the recursive on-line identification of the system model: optimized, adaptive and different fixed values of $\eta$ with 10 hidden neurons are used.**

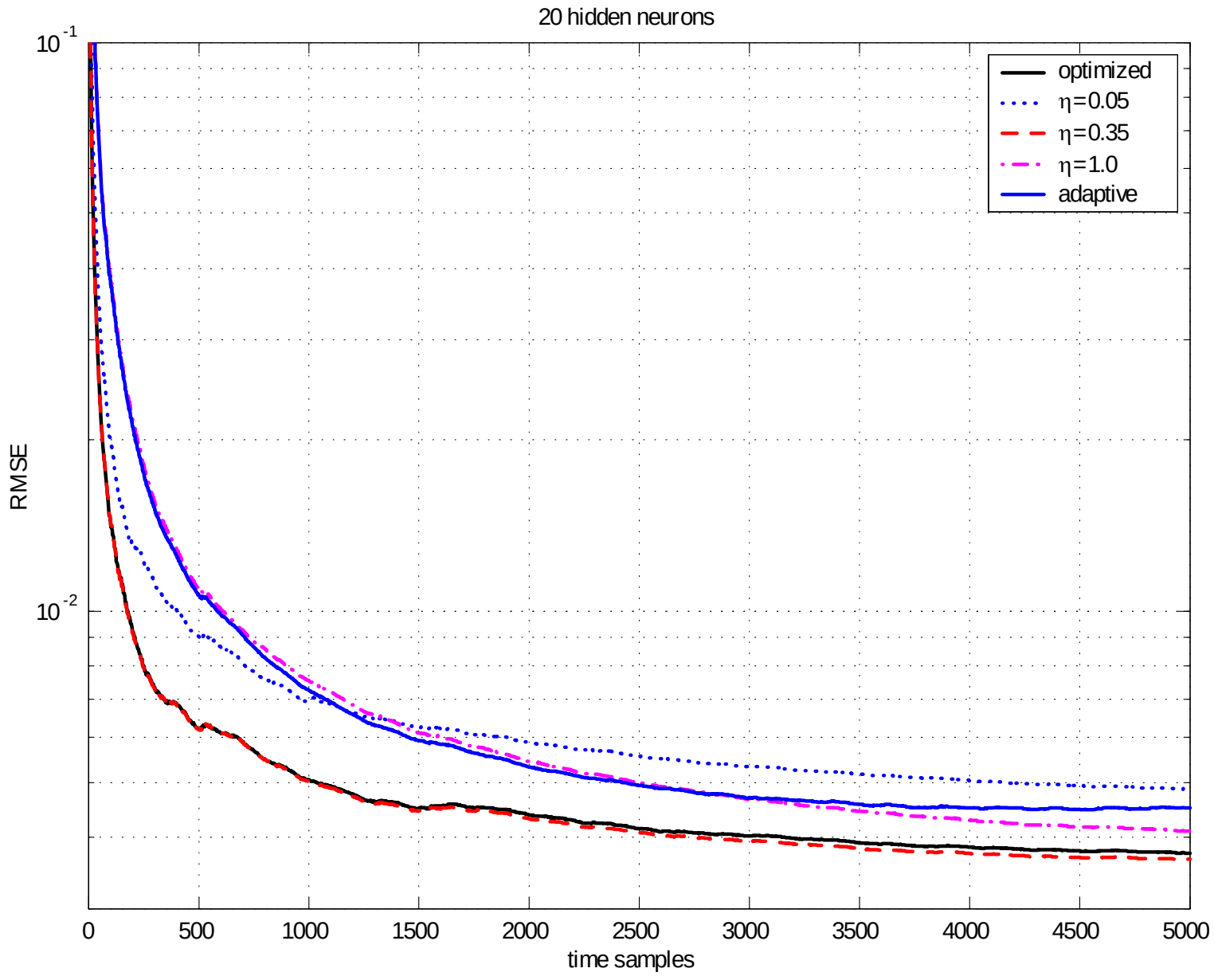

**Figure 3. RMSE in one-step-ahead prediction of system outputs based on the recursive on-line identification of the system model: optimized, adaptive and different fixed values of $\eta$ with 20 hidden neurons are used.**

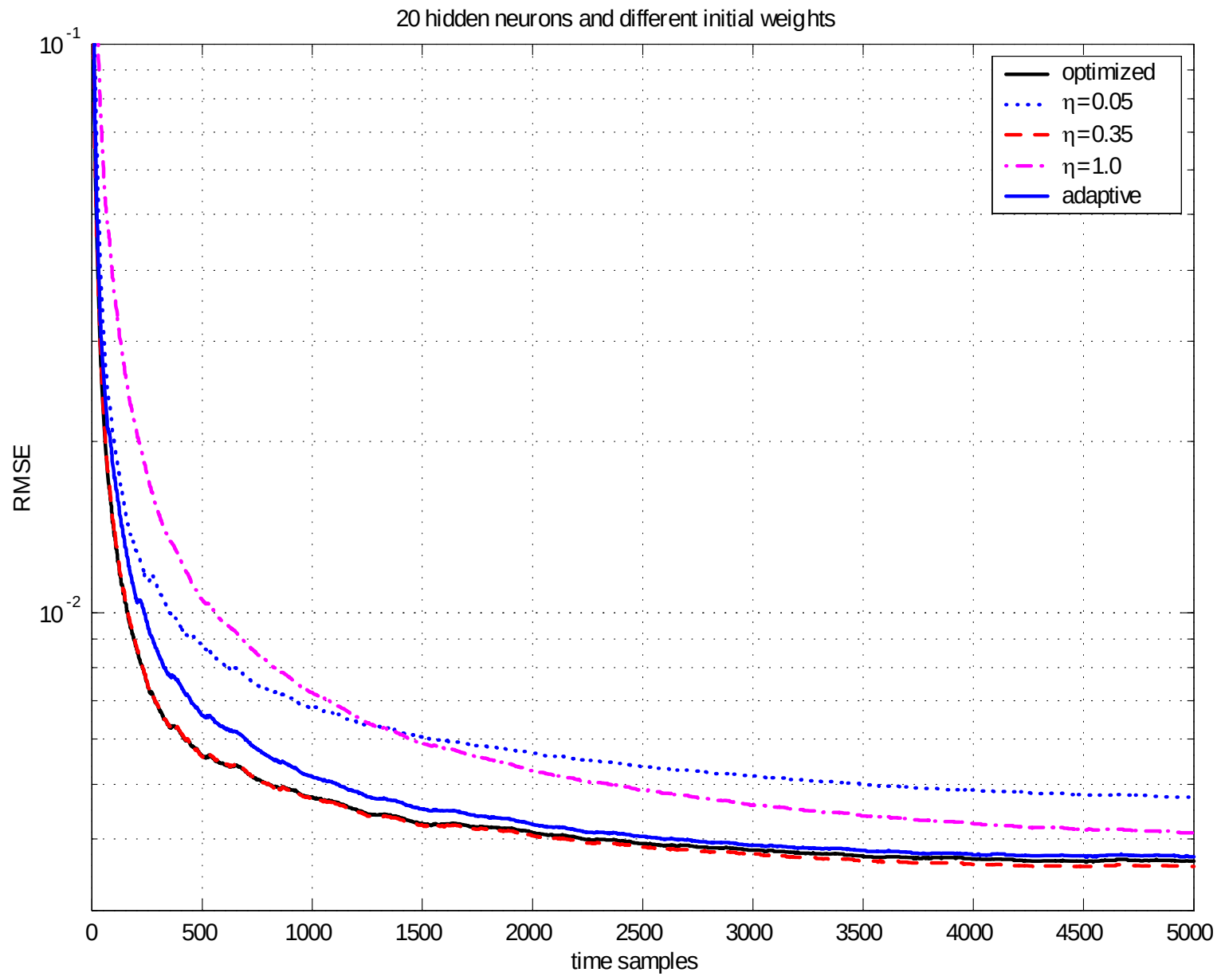

**Figure 4. RMSE in one-step-ahead prediction of system outputs based on the recursive on-line identification of the system model: optimized, adaptive and different fixed values of $\eta$ with 20 hidden neurons and different initial weights are used.**

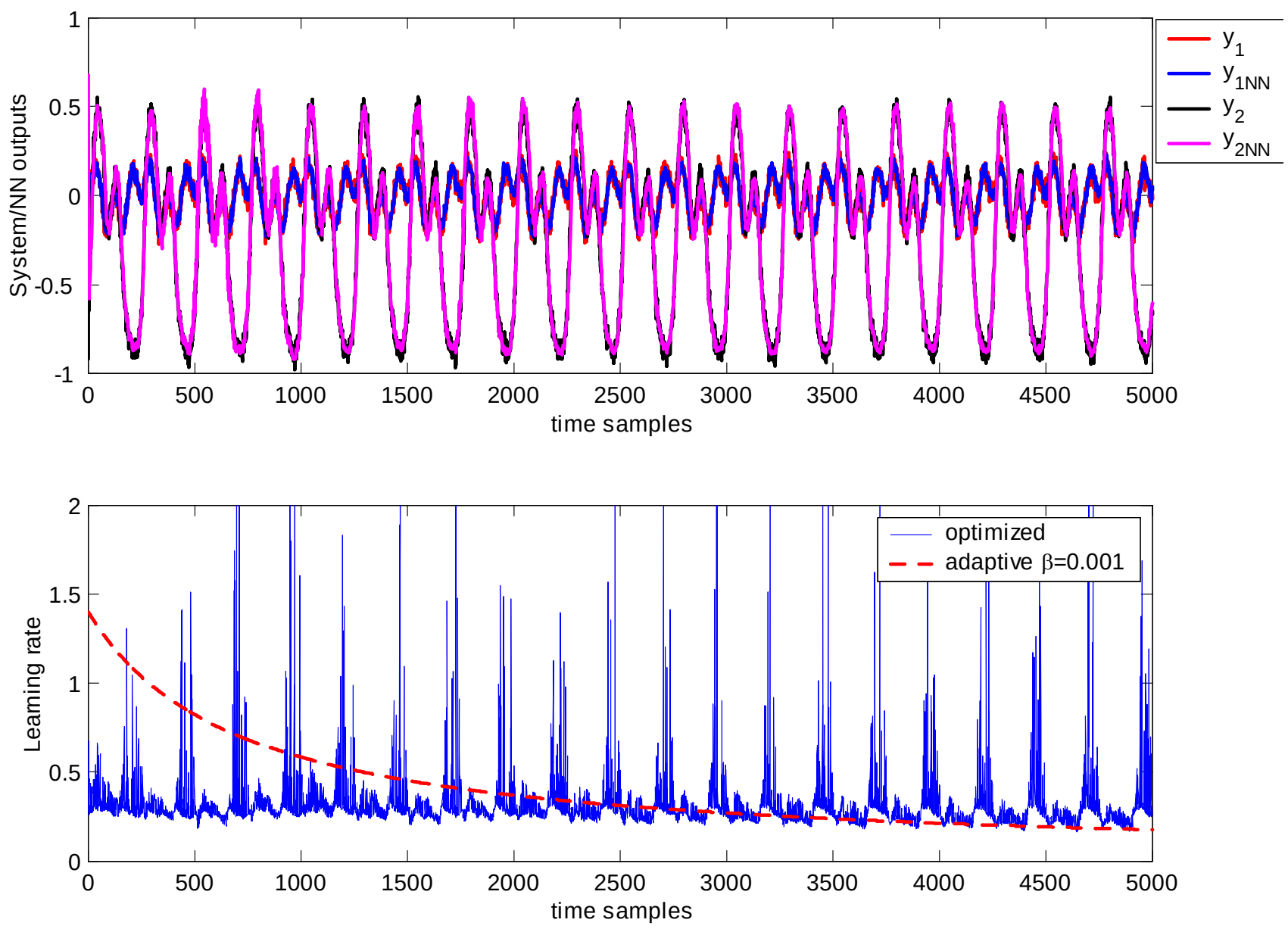

**Figure 5. A typical outputs of the system and neural networks and optimized and adaptive learning rate curves.**

## 5. CONCLUSIONS

This paper presents a new optimized variable learning rate via the matrix operation and optimization methods for recursive learning. The optimized learning algorithm always performs the best behavior near the optimal fixed learning rate, and eliminates the need for searching the proper fixed learning rate, and provides fast convergence. A weak convergence result, $E_2(t) \to E^*$ as $t \to \infty$ for the proposed algorithm indicates that the iteration procedure will converge to a local minimum of the error function $E^*$. The proposed algorithm is suitable for a multi-input and multi-output, three-layer (only one hidden layer), feed-forward neural networks being used for online identification of MIMO nonlinear systems. But it can be extended to multi-layer feed-forward neural networks. The effectiveness of the proposed algorithms applied to the identification of behavior of a two-input and two-output non-linear dynamic system is demonstrated by simulation experiments. The results achieved confirm the fast and stable learning of FNN.

## ABOUT THE AUTHORS

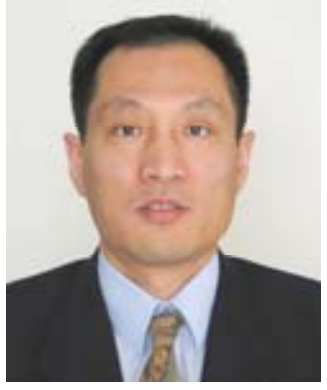

**D. Sha** received his PhD in Mechanical Engineering from Xi’an Jiaotong University, China in 1995. He is currently Biostatistician at Center for Clinical Epidemiology and Biostatistics, University of Pennsylvania School of Medicine. His previous research interests included applications of neural networks and fuzzy logic to control systems, modelling and simulation to electro-hydraulic systems and biomechanical systems.

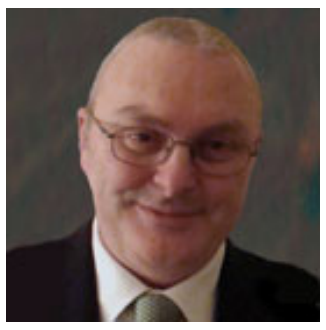

**V. B Bajic** earned a doctorate of Engineering Sciences in Electrical Engineering from the University of Zagreb in Croatia in 1989. He has been appointed Director of the Computational Bioscience Research Center and Professor of Applied Mathematics and Computational Science in the Mathematical and Computer Sciences and Engineering Division at KAUST, Saudi Arabia. Dr. Bajic was a Professor of Bioinformatics, and Acting and Deputy Director of the South African National Bioinformatics Institute (SANBI) at the University of the Western Cape.